\documentclass[letterpaper]{article} % DO NOT CHANGE THIS
\usepackage{aaai2026}
\usepackage{times}  % DO NOT CHANGE THIS
\usepackage{helvet}  % DO NOT CHANGE THIS
\usepackage{courier}  % DO NOT CHANGE THIS
\usepackage[hyphens]{url}  % DO NOT CHANGE THIS
\usepackage{graphicx} % DO NOT CHANGE THIS
\usepackage{amsmath}
\usepackage{natbib}  % DO NOT CHANGE THIS AND DO NOT ADD ANY OPTIONS TO IT
\usepackage{caption} % DO NOT CHANGE THIS AND DO NOT ADD ANY OPTIONS TO IT
\usepackage{algorithm}
\usepackage{algorithmic}
\usepackage{tabularx}

\usepackage{newfloat}
\usepackage{listings}
\usepackage{booktabs}
\usepackage{pifont}
\usepackage{amssymb}
\usepackage{listings}
\usepackage{xcolor}

\newcommand{\starfill}{\ding{72}}  % 实心星
\DeclareCaptionStyle{ruled}{labelfont=normalfont,labelsep=colon,strut=off} % DO NOT CHANGE THIS
\floatstyle{ruled}
\newfloat{listing}{tb}{lst}{}
\floatname{listing}{Listing}
\nocopyright % -- Your paper will not be published if you use this command
\title{EvoMakeup: High-Fidelity and Controllable Makeup Editing with MakeupQuad}

\author {
    Huadong Wu\thanks{Equal contribution.},
    Yi Fu\footnotemark[1],
    Yunhao Li,
    Yuan Gao,
    Kang Du
}
\affiliations {
    ByteDance \\
}

\usepackage{bibentry}
\usepackage{booktabs}
\begin{document}

\maketitle

\begin{abstract}
Facial makeup editing aims to realistically transfer makeup from a reference to a target face. Existing methods often produce low-quality results with coarse makeup details and struggle to preserve both identity and makeup fidelity, mainly due to the lack of structured paired data—where source and result share identity, and reference and result share identical makeup. To address this, we introduce MakeupQuad, a large-scale, high-quality dataset with non-makeup faces, references, edited results, and textual makeup descriptions. Building on this, we propose EvoMakeup, a unified training framework that mitigates image degradation during multi-stage distillation, enabling iterative improvement of both data and model quality. Although trained solely on synthetic data, EvoMakeup generalizes well and outperforms prior methods on real-world benchmarks. It supports high-fidelity, controllable, multi-task makeup editing—including full-face and partial reference-based editing, as well as text-driven makeup editing—within a single model. Experimental results demonstrate that our method achieves superior makeup fidelity and identity preservation, effectively balancing both aspects. Code and dataset will be released upon acceptance.
\end{abstract}

\begin{figure}[ht]
    \centering
    \includegraphics[width=\linewidth]{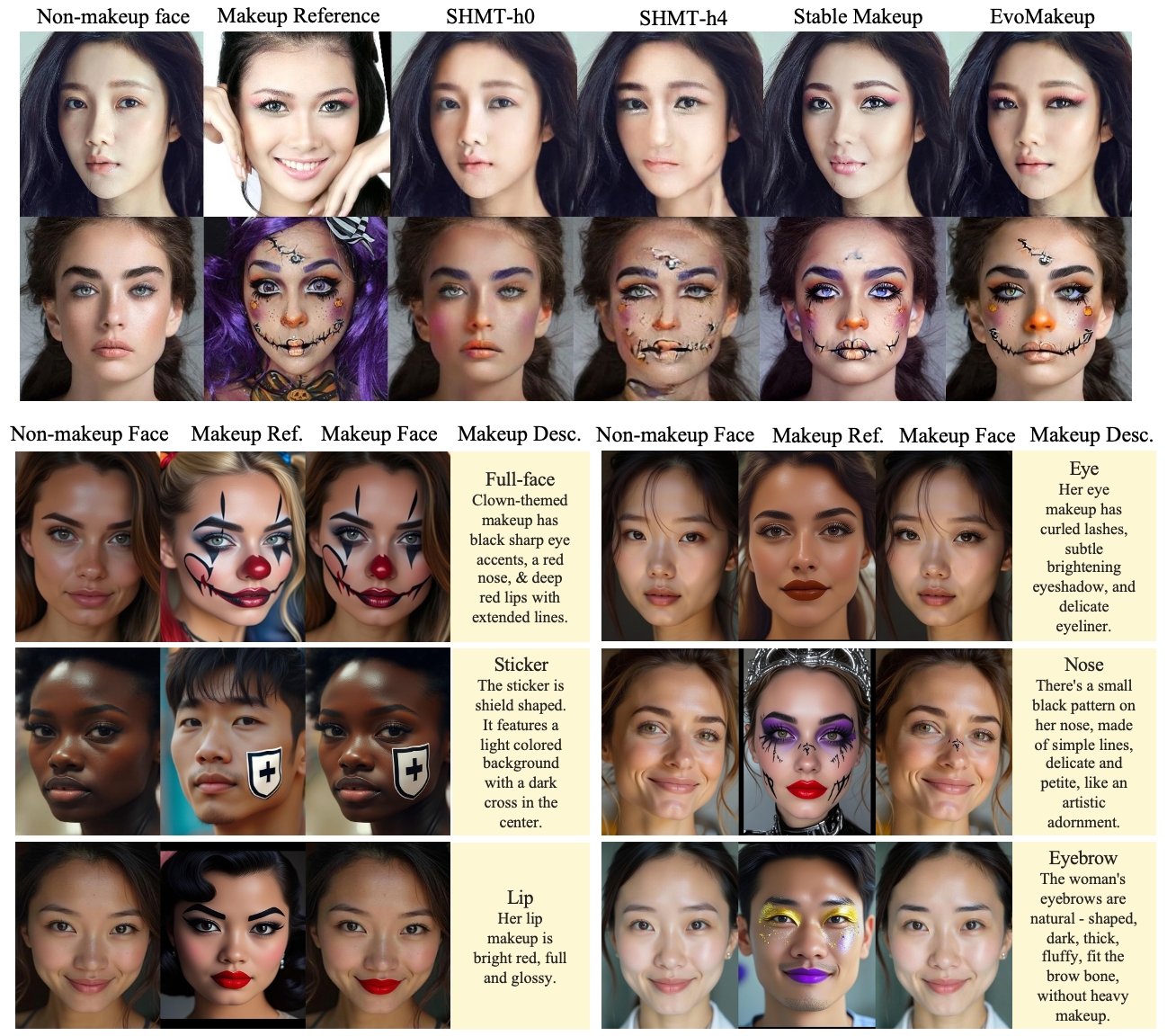}
    \caption{
    \textbf{Top:} Comparison with prior methods. EvoMakeup faithfully preserves both makeup and identity, while SHMT-h0, SHMT-h4, and Stable Makeup exhibit degraded makeup quality or identity loss.
    \textbf{Bottom:} MakeupQuad, the first quadruplet-based dataset, supports full-face and partial makeup editing with high identity and makeup consistency. It enables disentangled learning and supports EvoMakeup’s high-fidelity results.
    \textit{Please zoom in for more details.}
    }
    \label{fig:fig1}
\end{figure}

\section{Introduction}

Facial makeup editing modifies facial appearance by transferring or adjusting makeup. It plays a key role in entertainment, virtual try-on, recommendation, and AR. The task demands high-fidelity synthesis and precise control, posing challenges for image generation and localized editing. Despite recent progress, the task remains challenging.

Recent facial makeup editing methods include early GAN-based models (e.g., BeautyGAN~\cite{li2018beautygan}, PSGAN~\cite{jiang2020psgan}, SCGAN~\cite{zhao2020scgan}) and newer diffusion-based models (e.g., DiffAM~\cite{sun2024diffam}, SHMT~\cite{sun2024shmt}, Stable-Makeup~\cite{zhang2024stable}), which improve makeup realism and identity preservation. However, as shown in Fig.~\ref{fig:fig1} (top), neither early nor recent approaches fully succeed in simultaneously achieving high-fidelity preservation of both makeup and identity. Moreover, most methods suffer from limited makeup diversity, lack fine-grained partial editing—including SHMT and Stable-Makeup—and do not support multiple editing modes such as full-face, partial reference-based, and text-driven editing. These issues mainly stem from data and training limitations.

On the data side, existing methods are primarily trained on datasets such as \textit{Makeup-Wild}~\cite{jiang2020psgan}, \textit{LADN}~\cite{gu2019ladn}, and \textit{MT}~\cite{li2018beautygan}, which exhibit several limitations: (1) absence of high-quality triplets or quadruplets where source image \(I\) and makeup image \(M\) share the same identity, pose, and background, while makeup image \(M\) and reference \(R\) have identical makeup; (2) limited makeup diversity, predominantly featuring light or everyday styles, lacking complex or creative variations; and (3) low resolution and aesthetic quality, restricting applicability in real-world scenarios. The scarcity of paired data impedes effective disentanglement of identity and makeup, significantly constraining model performance. Furthermore, insufficient stylistic richness and suboptimal visual quality result in poor generalization and low-resolution outputs, hindering practical deployment. Given the substantial difficulty in collecting high-quality paired makeup data, few efforts have addressed large-scale dataset construction, leaving this area largely unexplored. Therefore, developing a high-quality, diverse, and scalable paired dataset remains both challenging and essential.

In terms of training paradigms, most models adopt single-stage supervised learning to extract and fuse identity and makeup features. SHMT encodes makeup using high-frequency signals, while Stable-Makeup employs pseudo-makeup features from augmented inputs. These methods mainly capture global styles and lack fine-grained regional control, limiting editing precision. Due to such constraints, no existing framework supports full-face, partial-region, and text-driven editing within a unified model.

To address these challenges, we construct MakeupQuad, the first high-quality quadruplet dataset for makeup editing. Automatically generated using advanced generative and editing models, it contains over 10,000 samples each with a source face, reference makeup image, edited result, and makeup description. As shown in Fig.~\ref{fig:fig1} (bottom), the dataset ensures strong consistency in makeup, identity, and image-text alignment, covering diverse makeup types including daily, light, heavy, and special-effect. MakeupQuad is the first to jointly support full-face, partial-region, and text-driven makeup editing.

Building upon this, we propose EvoMakeup, a unified framework trained via a model-data co-evolution strategy. By effectively addressing image degradation caused by multi-round distillation, EvoMakeup enables iterative refinement of both data quality and model performance. This process supports high-fidelity, highly controllable, and flexible makeup editing. Consequently, EvoMakeup significantly enhances identity preservation and makeup consistency, surpassing existing methods.

Our main contributions are summarized as follows:
\begin{itemize}
    \item We introduce and release MakeupQuad, the first large-scale, high-quality quadruplet dataset covering diverse makeup styles and supporting full-face, partial-region, and text-guided editing tasks.
    \item We propose EvoMakeup, a unified training framework that addresses editing degradation caused by multi-round distillation through data-model co-evolution, enabling high-fidelity, multi-task makeup editing within a single model.
    \item Extensive experiments on multiple benchmarks demonstrate that EvoMakeup outperforms previous state-of-the-art methods in identity preservation, makeup consistency, and image quality.
\end{itemize}

\begin{table*}[htbp]
\centering
\small
\begin{tabular}{lccccccccc}
\toprule
Dataset & No Makeup & Makeup & Ref. & Desc. & ID Pair & Makeup Pair & Quality & Diversity & Res. \\
\midrule

BeautyFace & 0 & 3,000 & 0 & 0 & \ding{55} & \ding{55} 
& \starfill\starfill\starfill & \starfill\starfill\starfill & 512 \\

MT Dataset & 1,115 & 2,719 & 0 & 0 & \ding{55} & \ding{55} &  
\starfill\starfill\starfill & 
\starfill\starfill\starfill & 361 \\

Wild Dataset & 335 & 384 & 0 & 0 & \ding{55} & \ding{55} & 
\starfill\starfill\starfill & 
\starfill\starfill\starfill & 256 \\

CPM Dataset & 1,625 & 1,625 & 1,625 & 0 & \ding{51} & \ding{51} & 
\starfill\starfill\starfill & 
\starfill\starfill\starfill & 256 \\

\textbf{MakeupQuad} & \textbf{10,000} & \textbf{10,000} & \textbf{10,000} & \textbf{10,000} & \ding{51} & \ding{51} & 
\starfill\starfill\starfill\starfill\starfill & 
\starfill\starfill\starfill\starfill\starfill & \textbf{$>$512} \\

\bottomrule
\end{tabular}
\caption{
Comparison of makeup datasets. Our \textbf{MakeupQuad} uniquely offers large-scale quadruplet annotations with paired identity and makeup references, delivering superior makeup quality, diversity, and higher resolution. 
Columns indicate counts of no-makeup images, makeup images, reference images, and makeup descriptions; presence of paired identity and makeup references; levels of makeup quality and diversity; and image resolution.
}
\label{tab:dataset_comparison}
\end{table*}

\section{Related Work}
\noindent\textbf{Makeup Editing Based on GANs and Diffusion Models.} Facial makeup editing has progressed from early GAN-based frameworks to diffusion-based methods. Classic GAN approaches, such as BeautyGAN~\cite{li2018beautygan}, PSGAN~\cite{jiang2020psgan}, and others~\cite{li2019asymmetric,deng2021spatially,hu2022protecting,kips2020gan,liu2021psgan++,wan2022facial,xiang2022ramgan,zhu2017unpaired,huang2021real,lyu2021sogan}, perform exemplar-based transfer with identity preservation but often face unstable training, limited style diversity, and weak fine-grained control. Later works integrate spatial priors—e.g., segmentation~\cite{gu2019ladn} and keypoints~\cite{nguyen2021lipstick}—and style-identity disentanglement~\cite{zhao2020scgan,yang2022elegant,xiang2022ramgan} for better controllability. Recent advances further employ attention~\cite{sun2022ssat,sun2024content} and invertible mappings~\cite{yan2023beautyrec} to enhance editing precision. Nonetheless, GAN-based models still struggle with localized realism, makeup diversity, and robustness.

Recent diffusion-based methods have significantly advanced makeup editing. DiffAM~\cite{sun2024diffam} introduces attribute-conditioned diffusion techniques to facilitate controllable makeup editing. More prominently, Stable-Makeup leverages CLIP-based supervision to enable both text- and image-driven makeup generation, achieving flexible and semantically meaningful editing. SHMT~\cite{sun2024shmt} enhances structural fidelity through hierarchical transformers, improving preservation of facial details and makeup consistency. MAD~\cite{ruan2025mad} supports multimodal inputs for fine-grained synthesis and manipulation guided by multiple cues. Despite these advances, diffusion models still face challenges in preserving fine local details such as eyelash textures and soft shadows, handling complex or personalized styles, and achieving natural blending in overlapping regions. Furthermore, their reliance on large annotated datasets is limited by the scarcity and low granularity of existing makeup data, restricting generalization and fine-level editing quality.

\noindent\textbf{Makeup Transfer Datasets.} High-quality and diverse datasets are essential for advancing makeup editing research. However, most existing datasets are unpaired, comprising before-and-after images from different identities, hindering the formation of strictly aligned samples. For example, \textit{BeautyFace}~\cite{yan2023beautyrec} includes only makeup-applied images; \textit{MT}~\cite{li2018beautygan} and \textit{Makeup-Wild}~\cite{jiang2020psgan} contain mismatched faces and makeup references, leading to reliance on suboptimal single-image training. \textit{CPM}~\cite{nguyen2021lipstick} introduces triplets of source, reference, and result, but its sticker-based synthetic construction yields low realism and limited quality.

Moreover, existing datasets generally lack support for localized editing and are limited to low resolutions (e.g., $256 \times 256$), restricting fine-grained, controllable, and realistic makeup transfer. While datasets such as \textit{LADN}~\cite{gu2019ladn}, \textit{MT-HR}~\cite{liu2021psgan++}, \textit{Wild-MT}~\cite{jiang2020psgan}, and \textit{MT-Text}~\cite{ruan2025mad} improve diversity, keypoint annotations, and textual descriptions, they still lack strongly paired samples with consistent identities and defined makeup regions. To address this, we construct the first quadruplet-based dataset, where each sample comprises a source face $I$, a reference image $R$, a result $M$, and a textual description $D$. This design improves training fidelity, enables precise regional control, and supports more generalizable, high-quality makeup generation.

\section{Method}
\label{sec:method}
\subsection{MakeupQuad Dataset}
\label{subsec:makeupquad}

High-quality paired datasets are vital yet challenging for facial makeup editing, requiring consistent identity, pose, and background between source and result images, and exact makeup matching between source and reference. Existing datasets have mismatches, low resolution, and limited diversity, hindering fine-grained editing.

To address these limitations, we introduce \textbf{MakeupQuad}, the first large-scale, high-resolution quadruplet dataset for makeup editing. Each quadruplet contains a source face \(I\), a makeup reference \(M\), the editing result \(R\), and a detailed makeup description \(D\) (see Fig.~\ref{fig:fig1} bottom), effectively disentangling identity and makeup information to improve identity preservation and makeup consistency. MakeupQuad supports full-face, partial-region, and instruction-driven editing, covering diverse styles, including daily, heavy, and special effects, and various ethnicities. As shown in Table~\ref{tab:dataset_comparison}, MakeupQuad surpasses prior datasets in scale, quality, and task diversity, providing a solid foundation for advancing makeup editing research.

\begin{figure*}[htbp]
    \centering
    \includegraphics[width=0.99\textwidth]{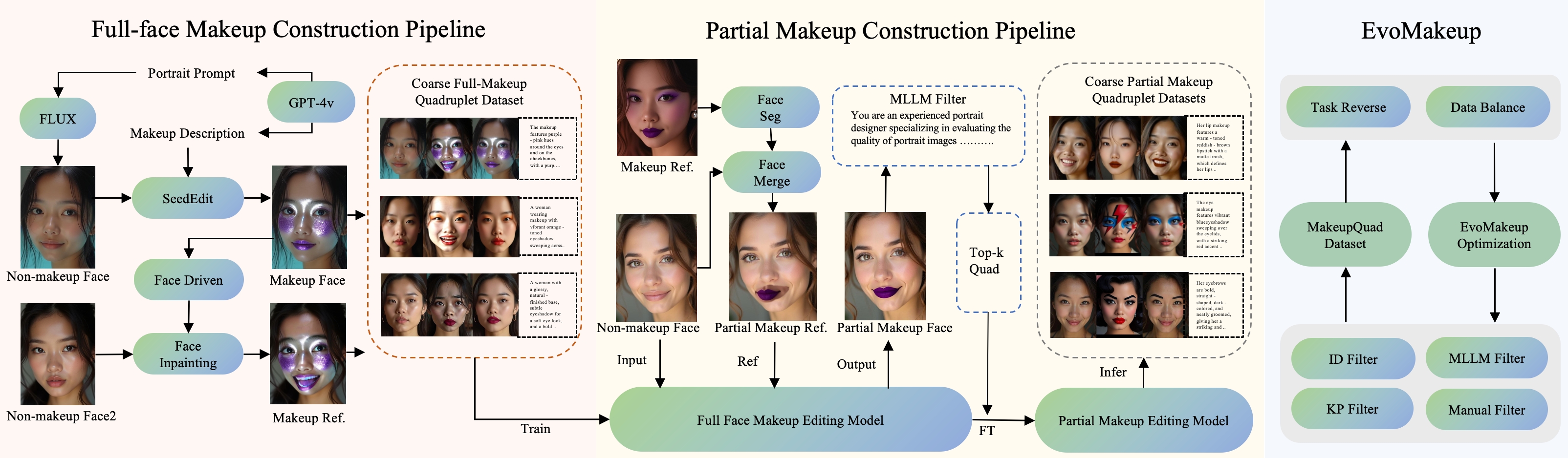}
    % \caption{
    % Overview of the \textbf{MakeupQuad} construction pipeline and the \textbf{EvoMakeup} training framework. 
    % \textbf{Left:} Initial full-face makeup quadruplets are constructed using GPT-4v-generated descriptions, identity-varied no-makeup images from FLUX, SeedEdit-based makeup editing, and inpainting to create reference images with consistent makeup but different identities. 
    % \textbf{Middle:} To build partial makeup data, a full-face editing model trained on coarse data is used to generate partial makeup results. High-quality region-specific tuples are selected via a MLLM-based filtering strategy and used to fine-tune partial editors, which then synthesize large-scale coarse quadruplets for partial makeup editing. 
    % \textbf{Right:} The EvoMakeup framework iteratively refines both model and data quality through co-evolution. To prevent degradation from repeated edits, the model-generated makeup quadruplets are reversed to form high-quality de-makeup data, which is injected into training as ground truth to stabilize learning and improve image fidelity.
    % Examples from the final MakeupQuad dataset are shown in the bottom part of Fig.~\ref{fig:fig1}.
    % }
    \caption{
    Overview of the \textbf{MakeupQuad} construction pipeline and the \textbf{EvoMakeup} training framework.  
    \textbf{Left:} Coarse full-face makeup quadruplets are built using GPT-4v-generated descriptions, identity-varied no-makeup images from FLUX, SeedEdit-based editing, and inpainting to create reference images with consistent makeup but different identities.  
    \textbf{Middle:} A full-face makeup editing model trained on the above coarse dataset generates partial makeup results. A limited set of high-quality region-specific quadruplets are selected through a series of filters, including an MLLM-based strategy, to fine-tune partial editing models that synthesize large-scale coarse partial quadruplets.  
    \textbf{Right:} The EvoMakeup framework iteratively refines both model and data based on these coarse datasets. Using task reverse and data balance strategies with multiple filtering steps, EvoMakeup progressively improves fidelity and consistency. The final MakeupQuad dataset produced by this co-evolution is shown at the bottom of Fig.~\ref{fig:fig1}.
    }
    
    \label{fig:pipeline}
\end{figure*}

\subsection{Full-Face Makeup Data Construction}
\label{subsec:fullface}

The construction of full-face makeup data forms the basis of the MakeupQuad dataset. As shown in the left of Fig.~\ref{fig:pipeline}, this process involves the following steps:

\noindent\textbf{Non-Makeup Faces \(I\).} We begin by using GPT-4v~\cite{hurst2024gpt} to generate a diverse set of portrait-related prompts. These prompts are fed into FLUX~\cite{flux2024} to synthesize a large number of high-quality, makeup-free facial images. The resulting dataset exhibits wide variation in ethnicity and gender.

\noindent\textbf{Makeup Faces \(M\).} To generate makeup versions of these faces, we use SeedEdit~\cite{wang2025seededit}, a state-of-the-art editing model known for strong identity preservation and structural consistency. We first create ~1,000 diverse makeup prompts via GPT-4v, covering natural, heavy, and special effects makeup styles. These prompts are applied to the non-makeup faces \(I\) using SeedEdit. We filter the outputs based on facial alignment, identity similarity, and makeup realism. Finally, we use facial masks to blend the edited face back into the original image, restricting changes to the facial region and preserving the background.

\noindent\textbf{Makeup Reference Images \(R\).}  
A key challenge in our pipeline is constructing suitable makeup references that (1) differ in identity from both the non-makeup image \(I\) and the makeup-applied image \(M\), and (2) preserve the makeup appearance of \(M\). Prior methods (e.g., Stable-Makeup) relying on geometric perturbations often fail to achieve this balance, either preserving identity or distorting makeup.

To address this, we introduce a reference generation pipeline that first applies LivePortrait~\cite{guo2024liveportrait} to animate \(M\), producing \(M'\) with altered facial structure while preserving makeup details. We then use MimicBrush~\cite{chen2024zero} to transfer \(M'\) onto the facial layout of a different non-makeup face \(I'\), guided by its mask, generating cross-identity reference candidates \(R\). Here, \(I'\) is sampled from a separate identity pool and shares no identity with \(I\), \(M\), or \(M'\), ensuring identity disentanglement. A filtering step selects \(R\) images that are visually consistent and identity-independent. This pipeline overcomes limitations of prior methods and MimicBrush alone, whose outputs often suffer identity leakage or makeup degradation.

\noindent\textbf{Makeup Description \(D\).}  
As mentioned, we use GPT-4v to generate about 1,000 diverse makeup prompts to guide SeedEdit in synthesizing makeup-applied images. However, due to prompt-image alignment limits, these initial descriptions may not fully reflect the actual makeup details in \(M\). To address this, we use GPT-4v to refine annotations for each image, producing the final makeup description \(D\).

In summary, we construct a coarse full-face makeup quadruplet dataset \(\{I, M, R, D\}\), where \(I\) is the non-makeup face, \(M\) the makeup-applied image, \(R\) the reference, and \(D\) the detailed makeup description. We derive subsets for tasks including instruction-based makeup application \(\{I, M, D\}\) and makeup removal \(\{M, I, D'\}\), with \(D'\) representing removal instructions. The following sections describe how we further decompose facial regions to support localized makeup tasks and improve data quality.

\subsection{Partial Makeup Data Construction}
\label{subsec:partial}

In real-world scenarios, makeup editing often involves partial modifications, such as adjusting lip color or trying new eyeshadow, rather than full-face makeup editing. While models trained on full-face data can perform global makeup editing effectively, they generally lack the precision and controllability required for fine-grained partial editing. This motivates the need for dedicated partial makeup data to support high-fidelity, region-specific editing.

Constructing such data is more challenging than building full-face datasets. Although general editing models generate partial makeup effects, they often suffer from limited spatial control: edited regions may not align with instructions, and unedited areas can be unintentionally altered in structure or texture. Therefore, the global makeup editing pipeline described in Sec.3.2 is unsuitable for partial editing tasks.

To address this, we propose a \textit{model-based} pipeline leveraging our full-face dataset and pretrained model to generate coarse region-specific makeup data, as briefly summarized in the middle of Fig.~\ref{fig:pipeline}.

\noindent\textbf{Training a Full-Face Makeup Model.}  
We fine-tune a pretrained OmniGen~\cite{xiao2025omnigen} checkpoint—an open-source diffusion model with strong generative and representational capabilities—on our full-face dataset, updating all parameters except the VAE module. The resulting model effectively transfers makeup from references while disentangling identity and makeup features. Serving as the foundation for localized data generation, it applies makeup from partially made-up references to corresponding facial regions on source images, enabling high-quality, region-specific makeup synthesis.

\noindent\textbf{Generating Initial Partial Makeup Data.}  
We first generate coarse partial makeup references \(R'\) by compositing masked regions—such as eyes, eyebrows, lips, nose, and stickers—from a full-face makeup reference image \(R\) onto a non-makeup face \(I\). Masks are obtained via Mediapipe~\cite{lugaresi2019mediapipe} and Grounded SAM~\cite{ren2024grounded}. Then, leveraging the pretrained full-face makeup model, we apply makeup from \(R'\) onto \(I\) to synthesize partial makeup images \(M\). This process yields a large but noisy set of partial makeup results. To ensure quality, we use GPT-4v to filter and retain only samples with high visual consistency, structural fidelity, and editing precision. The resulting triples \(\{I, R, M\}\) serve as high-quality partial makeup training data for subsequent model fine-tuning.

\noindent\textbf{Training LoRA Modules for Each Makeup Type.}  
For each partial makeup category, we use the filtered high-quality samples to train lightweight LoRA modules on top of the full-face makeup model. Leveraging the base model’s prior training in global editing—which provides strong identity-makeup disentanglement and reference-based editing capabilities—these LoRA modules require only a few samples to specialize in specific partial makeup (e.g., lipstick, eyeshadow), while preserving identity and ensuring partial consistency with the reference.

\noindent\textbf{Generating Partial Makeup Data at Scale.}  
Each trained LoRA module is used to batch-generate partial makeup samples \(\{I, R, M\}\), where \(I\) is the non-makeup face, \(R\) the full-face makeup reference, and \(M\) the partially edited result. GPT-4V annotates each output with a detailed description \(D\) of the applied makeup.

In summary, this coarse dataset covers five partial makeup categories: \textit{eye}, \textit{lip}, \textit{eyebrow}, \textit{nose}, and \textit{sticker}. Each sample is a quadruplet \(\{I, R, M, D\}\), containing the non-makeup face \(I\), full-face reference \(R\), partially edited image \(M\), and corresponding makeup description \(D\).

\begin{table*}[htbp]
    \centering
    \begin{tabular}{l  c  c  c  c  c  c  c  c  c  c  c  c}
        \toprule
            & \multicolumn{4}{c}{MT Testset} & 
              \multicolumn{4}{c}{Wild-MT Testset} & 
              \multicolumn{4}{c}{LADN Testset}\\
        \midrule
        Method  & FID↓ & FSim↑ & MSim↑ & NEP↑
                & FID↓ & FSim↑ & MSim↑ & NEP↑
                & FID↓ & FSim↑ & MSim↑ & NEP↑ \\
        \midrule

        PSGAN
                         & 70.3 & 0.80 & 3.03  & 9.52
                         & 76.6 & 0.82 & 2.37  & 9.66 
                         & 54.2 & 0.81 & 2.70  & 9.70 \\
        EleGANt
                         & 78.7  & 0.82 & 3.18 & 9.68     
                         & 101.3 & 0.82 & 2.55 & 9.58       
                         & 53.1  & 0.81 & 3.35 & 9.68  \\
        SSAT
                         & 73.6 & 0.70 & 2.60 & 7.51  
                         & 88.7 & 0.76 & 2.22 & 8.49 
                         & 59.0 & 0.72 & 2.47 & 8.02  \\
        StableMakeup
                         & 62.5 & 0.48 & 4.51 & 9.74  
                         & 67.0 & 0.48 & 4.83 & 9.79 
                         & 35.4 & 0.47 & 4.78 & 9.71  \\
        SHMT-h0
                         & 75.5 & \textbf{0.86} & 3.42 & 9.91 
                         & 75.7 & \textbf{0.88} & 2.65 & 9.91 
                         & 49.3 & \textbf{0.84} & 3.46 & 9.90  \\
        SHMT-h4
                         & 63.3 & 0.27 & 3.85 & 9.52  
                         & 65.8 & 0.32 & 3.10 & 9.75 
                         & 36.0 & 0.24 & 4.56 & 9.55  \\
        MAD
                         & 105.7 & 0.71 & 1.37 & 9.70  
                         & 125.3 & 0.60 & 1.65 & 9.85 
                         & 102.2 & 0.63 & 1.21 & 9.81  \\
        \textbf{EvoMakeup}
                         & \textbf{57.5} & 0.73 & \textbf{6.21} & \textbf{9.94}  
                         & \textbf{63.8} & 0.66 & \textbf{6.46} & \textbf{9.94} 
                         & \textbf{32.9} & 0.65 & \textbf{6.30} & \textbf{9.91}  \\
        \bottomrule
    \end{tabular}
    \caption{
    Quantitative comparison of \textbf{EvoMakeup} and prior methods on MT, Wild-MT, and LADN.  Lower FID and higher FSim (Identity), MSim (Makeup), and NEP (Non-Edited Region Preservation) indicate better performance.
    }
    \label{tab:quantitative}
\end{table*}

\begin{table*}[htbp]
    \centering
    \begin{tabular}{l  c  c  c  c  c  c  c  c  c  c  c  c}
        \toprule
            & \multicolumn{4}{c}{MT Testset} & 
              \multicolumn{4}{c}{Wild-MT Testset} & 
              \multicolumn{4}{c}{LADN Testset}\\
        \midrule
        Method & Makeup & ID & NEP & Quality
               & Makeup & ID & NEP & Quality
               & Makeup & ID & NEP & Quality \\
        \midrule

        PSGAN
                         & 0.0 & 17.2 &  2.8 & 0.0 
                         & 0.0 & 17.0 &  7.2 & 0.0 
                         & 0.0 & 12.3 & 15.1 & 0.0 \\
        EleGANt
                         & 4.8 & 22.4  & 3.6 & 2.4      
                         & 1.2 & 25.0  & 2.0 & 2.7      
                         & 2.0 & 27.4  & 9.2 & 2.4   \\
        SSAT
                         & 0.0 & 0.0 & 0.4 & 0.0
                         & 0.0 & 0.4 & 0.4 & 0.0
                         & 0.0 & 0.0 & 0.5 & 0.0 \\
        StableMakeup
                         & 18.4 & 0.0 & 4.7 & 26.0 
                         & 17.6 & 0.8 & 0.8 & 34.5
                         & 16.8 & 1.6 & 1.1 & 32.0 \\
        SHMT-h0
                         & 0.8 & \textbf{36.1} & 24.6 & 2.0 
                         & 0.0 & \textbf{28.4} & 36.9 & 1.2
                         & 0.0 & \textbf{31.4} & 32.2 & 3.4 \\
        SHMT-h4
                         & 14.8 & 0.0 & 12.2 & 6.0 
                         &  4.0 & 0.0 &  8.8 & 0.5
                         &  9.2 & 0.0 &  2.1 & 2.0 \\
        MAD
                         &  0.0 & 5.6 &  3.1 & 0.0 
                         &  0.0 & 3.0 &  1.1 & 0.0
                         &  0.0 & 1.2 &  2.0 & 0.0 \\
        \textbf{EvoMakeup}
                         & \textbf{61.2} & 18.7 & \textbf{48.6} & \textbf{63.6} 
                         & \textbf{77.2} & 25.4 & \textbf{42.8} & \textbf{61.1}
                         & \textbf{72.0} & 26.1 & \textbf{37.8} & \textbf{60.2} \\
        \bottomrule
    \end{tabular}
    \caption{
    User study results on MT, Wild-MT, and LADN. Values denote top-1 vote percentages. Higher is better for all metrics: Makeup consistency, identity preservation (ID), non-edited region preservation (NEP), and overall quality.
    }
    \label{tab:userstudy}
\end{table*}

\subsection{EvoMakeup: A Co-Evolution Framework}
\label{subsec:evomakeup}

Although our coarse MakeupQuad dataset supports diverse makeup editing tasks—including full-face and partial reference-based editing as well as instruction-driven editing—its construction partially relies on model-generated pseudo-labels that suffer from issues such as insufficient identity similarity, inconsistent makeup, and unintended modifications in non-editing regions.

To improve both pseudo-label dataset quality and model performance, we propose \textbf{EvoMakeup}, a co-evolution framework that iteratively optimizes the editing model and dataset, as illustrated on the right of Fig.~\ref{fig:pipeline}. At each iteration, the model generates new pseudo-labeled data from the current dataset, which are then filtered by a quality-aware module using keypoint alignment, identity similarity, MLLM evaluation, and manual selection. 

A critical challenge in this iterative process is preventing data quality degradation—manifested as blurring, structural distortions, and color shifts—that can accumulate and destabilize training. To address this, we introduce a novel dynamic data composition strategy that anchors training with high-quality samples drawn from distilled and filtered task-reversed data (i.e., makeup removal), using undistilled non-makeup faces as ground truth. By carefully balancing these with other samples, our approach maintains strong identity and makeup consistency, improves overall data fidelity, and ensures stable model convergence.

This co-optimization process enables us to construct a high-quality MakeupQuad dataset with enhanced realism, fine-grained controllability. Built on this dataset, \textbf{EvoMakeup}, leveraging the OmniGen diffusion architecture, supports unified text- and image-guided editing, achieving high-fidelity fine-level makeup transfer with robust generalization across diverse tasks.

\begin{figure*}[ht]
    \centering
    \includegraphics[width=0.9\textwidth]{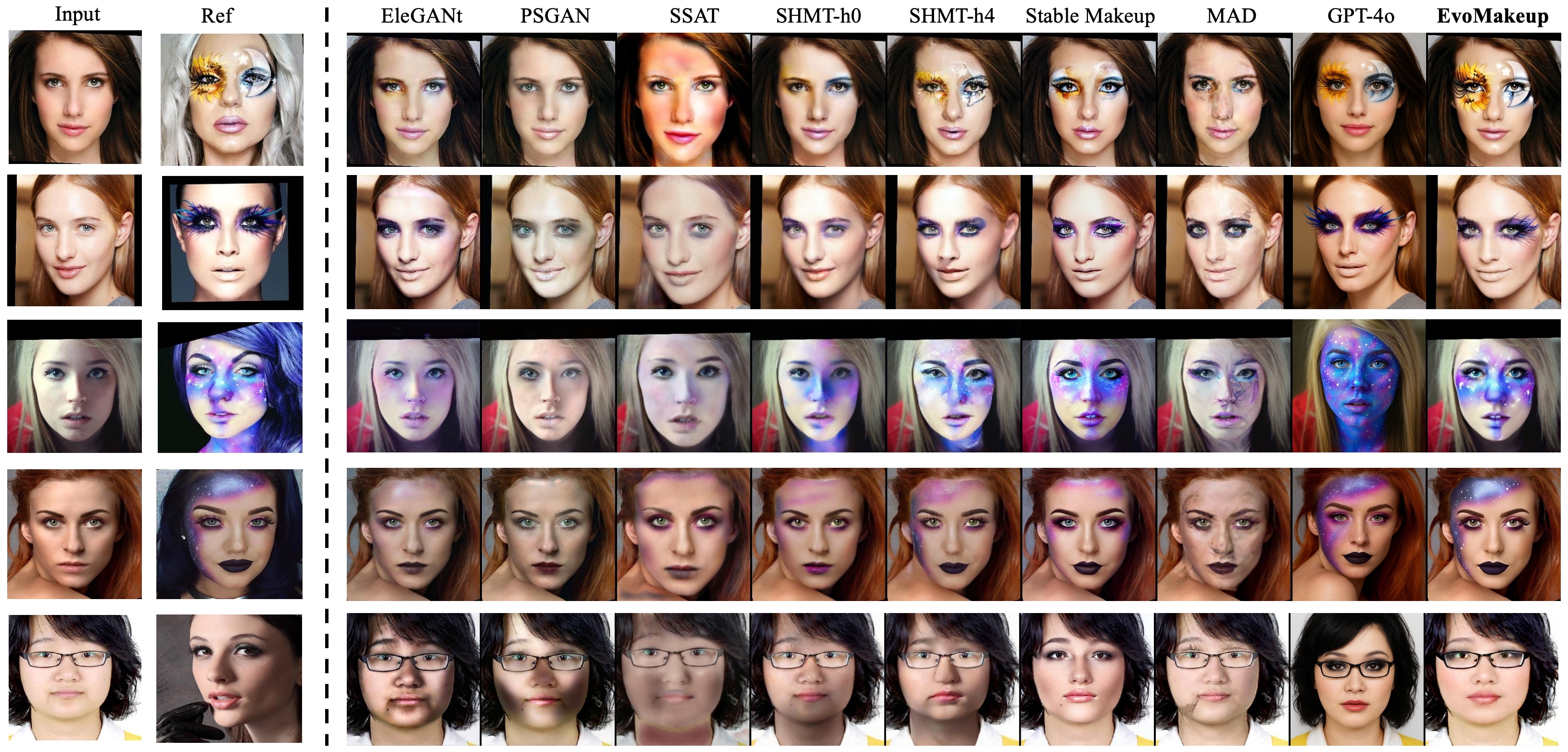}
    \caption{Qualitative comparison of EvoMakeup with GAN and diffusion approaches for full-face makeup editing.}
    \label{fig:qualitative}
\end{figure*}

\begin{figure*}[ht]
    \centering
    \includegraphics[width=0.9\linewidth]{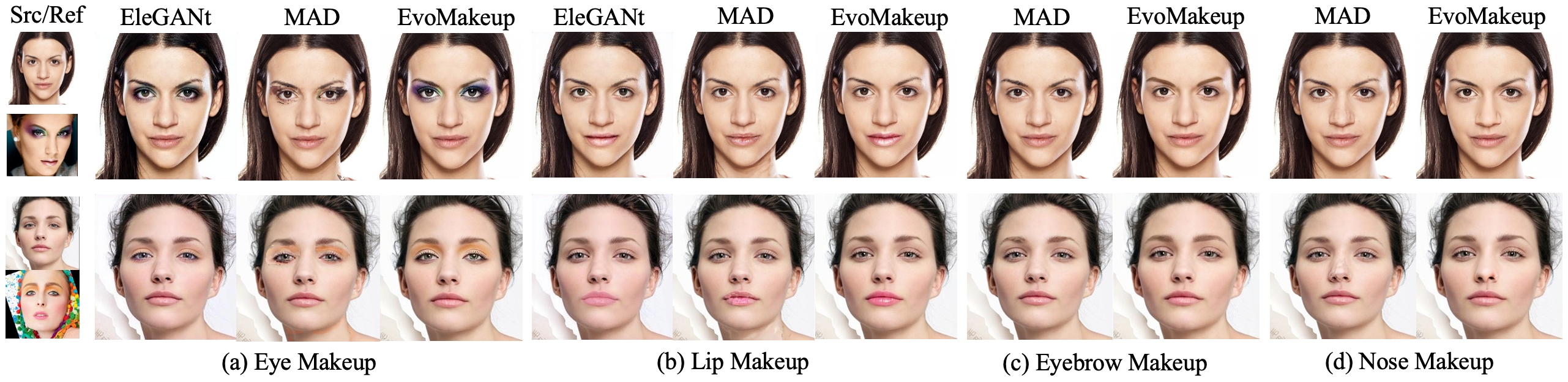}
    \caption{Qualitative comparison with GAN/diffusion methods for partial edits; EleGANt lacks eyebrow/nose support.
    }
    \label{fig:partial}
\end{figure*}

\section{Experiments}

We conduct comprehensive experiments to evaluate the effectiveness of our method in terms of identity preservation, makeup consistency, editability, and overall image quality.

\subsection{Experimental Settings}

\noindent\textbf{Training Details.}  
We use OmniGen as the backbone, initialized from its public checkpoint. 
We first fine-tune all parameters (except the VAE) on coarse full-face dataset to train the \textbf{full-face makeup model} (Sec.~3.3), using AdamW (lr = 1e-4) for 20K steps with a batch size of 8 on 8$\times$A800 GPUs. 
\textbf{Partial-makeup models} are trained on top-ranked region-specific samples using LoRA adapters (rank 8) for 1K steps with lr = 3e-4.
We then build the \textbf{EvoMakeup model} through two rounds of model-data co-evolution on the coarse dataset, each fine-tuning all parameters (except the VAE) for 50K steps with the same setup.
The final EvoMakeup model achieves strong generalization and high-quality editing through iterative data refinement.

\begin{table*}[htbp]
    \centering
    \begin{tabular}{l  c  c  c  c  c  c  c  c  c c c c}
        \toprule
            & \multicolumn{4}{c}{MT Testset} & 
              \multicolumn{4}{c}{Wild-MT Testset} & 
              \multicolumn{4}{c}{LADN Testset}\\
        \midrule
        Method & FID$\downarrow$ & FSim$\uparrow$ & MSim$\uparrow$ & NEP$\uparrow$
        & FID$\downarrow$ & FSim$\uparrow$ & MSim$\uparrow$ & NEP$\uparrow$
        & FID$\downarrow$ & FSim$\uparrow$ & MSim$\uparrow$ & NEP$\uparrow$ \\
        \midrule
        EvoMakeup$^{0}$
                         & 58.3 & 0.41 & 4.53 & 9.79
                         & 73.1 & 0.37 & 5.02 & 9.77
                         & 37.8 & 0.36 & 5.53 & 9.76 \\
        EvoMakeup$^{1}$
                         & 57.9 & 0.72 & 5.94 & 9.83 
                         & 71.8 & 0.62 & 5.97 & 9.84
                         & 33.6 & 0.63 & 6.11 & 9.88 \\
        EvoMakeup$^{2}$
                         & \textbf{57.5} & \textbf{0.73} & \textbf{6.21} & \textbf{9.94} 
                         & \textbf{63.8} & \textbf{0.66} & \textbf{6.46} & \textbf{9.94}
                         & \textbf{32.9} & \textbf{0.65} & \textbf{6.30} & \textbf{9.91} \\
        \bottomrule
    \end{tabular}
    \caption{
    Ablation study of \textbf{EvoMakeup} across data-model co-evolution iterations. EvoMakeup$^{0}$ is the initial model; EvoMakeup$^{1}$ and $^{2}$ show improvements after first and second evolutions. Performance improves consistently across all metrics.
    }
    \label{tab:ablation}
\end{table*}

\noindent\textbf{Evaluation Datasets.}
We construct three test sets by randomly selecting 1000 pairs from each of \textit{MT}~\cite{li2018beautygan}, \textit{Wild-MT}~\cite{jiang2020psgan}, and \textit{LADN}~\cite{gu2019ladn}. \textit{Wild-MT} includes significant pose and expression variations, while \textit{LADN} features diverse and complex makeup styles. In total, our evaluation covers 3,000 image pairs.

\noindent\textbf{Evaluation Metrics.}
To objectively compare methods, we adopt these metrics. FID~\cite{heusel2017gans} measures realism by comparing generated images with reference makeup images (lower is better). Identity preservation is evaluated via facial similarity~\cite{deng2021insightface} between input and edited images. Makeup consistency and non-edited region preservation (NEP) are assessed using GPT-4v-based metrics (details in Supplementary).

We also conduct a user study with side-by-side full-face results in random order. Twenty participants select the best image based on makeup consistency, ID preservation, NEP, and overall quality. We report top-1 vote percentages.

\subsection{Comparison with State-of-the-Art Methods}

\noindent\textbf{Baselines.}  
We compare six recent methods using official code and models: PSGAN, EleGANt, SSAT (GAN); Stable-Makeup, SHMT, MAD (diffusion).

\noindent\textbf{Quantitative Results.}  
We evaluate all methods on the MT, Wild-MT, and LADN test sets. Table~\ref{tab:quantitative} reports four metrics: FID (lower is better), facial identity similarity (FSim), makeup similarity (MSim), and non-edited region preservation (NEP) (higher is better). Our proposed \textbf{EvoMakeup} achieves the best or competitive results across all benchmarks, notably excelling in MSim and NEP, indicating superior makeup editing quality and background preservation.

A common limitation of existing methods is the trade-off between identity preservation and realistic makeup editing: approaches with high FSim (e.g., SHMT-h0) often exhibit poor MSim, while those with strong makeup transfer (e.g., StableMakeup) tend to compromise identity. Notably, even methods with low FSim frequently fail to preserve consistent makeup, suggesting intrinsic limitations in disentangling identity and makeup. In contrast, EvoMakeup is the only method that simultaneously achieves high identity preservation and high makeup fidelity, demonstrating its effectiveness in resolving this long-standing conflict. Overall, EvoMakeup offers the best trade-off among all methods.

\noindent\textbf{User Study.}
To complement quantitative evaluation, we conduct a user study where participants compare full-face makeup editing results from various methods. Table~\ref{tab:userstudy} summarizes top-1 vote percentages for makeup consistency, identity preservation (ID), non-edited region preservation (NEP), and overall quality. While some methods score higher in identity preservation due to weaker makeup transfer, EvoMakeup consistently ranks best in makeup consistency, NEP, and overall quality, demonstrating its advantage in producing visually appealing, identity-consistent, and regionally coherent makeup results.

\noindent\textbf{Qualitative Results.}  
Fig.~\ref{fig:qualitative} shows qualitative comparisons of full-face makeup editing between our method and representative GAN- and diffusion-based approaches, including simple and complex makeup. Our method effectively transfers fine details such as feathered eyeshadow and hair strands, and handles vibrant, large-area makeup robustly, even with occlusions like eyeglasses. Elegant, PSGAN, SSAT, and SHMT-h0 show inferior makeup quality, while StableMakeup, SHMT-h4, and GPT-4o exhibit weaker identity preservation and less faithful makeup transfer. Overall, our approach excels in editing fidelity, identity preservation, non-edited region consistency, and image quality.

Fig.~\ref{fig:partial} presents qualitative comparisons of partial makeup editing—including eye, lip, eyebrow, and nose—between our method and two baselines: EleGANt (GAN-based) and MAD (diffusion-based).  
Our approach achieves more precise and faithful local makeup transfer. For eye makeup, it preserves fine textures and renders more complete, natural results. Lip makeup accurately transfers color and gloss, yielding coherent and appealing effects. In the eyebrow region, where other methods show minimal or no changes, our method maintains brow bone structure while adapting shape, color, and individual hairs. For nose contouring, it effectively reproduces shading and reduces local skin imperfections—capabilities largely missing in the baselines.

\subsection{Ablation Study}
\noindent\textbf{Effectiveness of the EvoMakeup Framework.}  
The EvoMakeup training framework adopts a multi-round data-model co-optimization approach to progressively enhance the identity and makeup consistency of both the dataset and the model, while maintaining image quality. Specifically, as shown in Table~\ref{tab:ablation}, EvoMakeup$^{0}$ represents the baseline trained on the original dataset; EvoMakeup$^{1}$ denotes the model after one iteration; and EvoMakeup$^{2}$ is the final model trained on the MakeupQuad dataset. EvoMakeup$^{2}$ consistently achieves higher facial similarity and makeup consistency scores, demonstrating the effectiveness of the iterative strategy. Moreover, its lower FID score indicates improved visual quality and reduced degradation throughout the distillation process.

\begin{figure}[ht]
    \centering
    \includegraphics[width=0.83\linewidth]{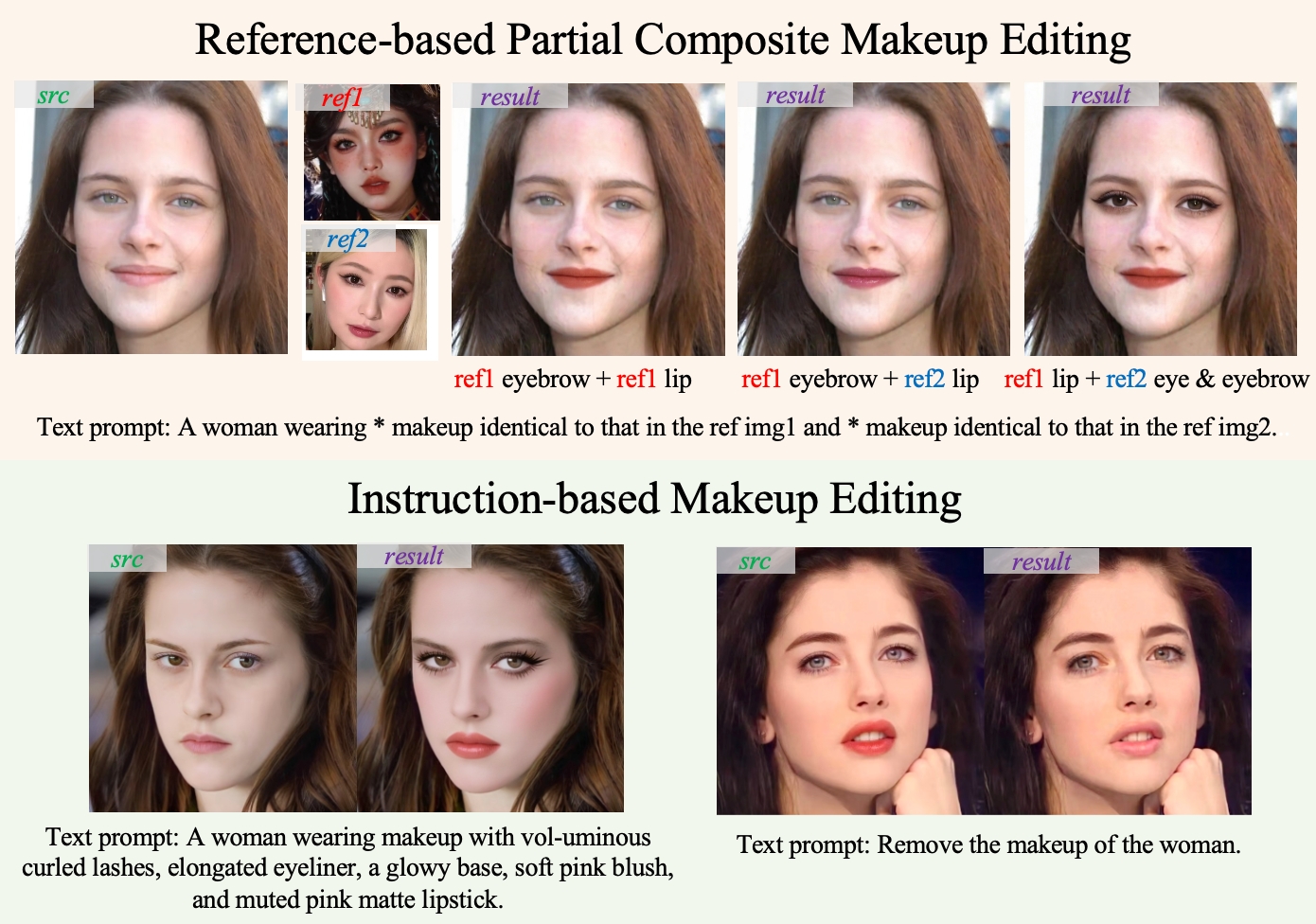}
    \caption{Additional applications of EvoMakeup.}
    \label{fig:more}
\end{figure}

\subsection{Additional Applications}
As shown in Fig.~5, our method supports full-face and partial makeup editing guided by reference images, as well as compositional editing using multiple regions from one or more references. It also enables instruction-based makeup editing, including application and removal via text. These results demonstrate EvoMakeup’s high fidelity, controllability, and versatility. Our method further supports cross-domain editing, with inputs ranging from real to stylized faces, and references including real, synthetic, or even abstract images. More results are provided in the supplement.

\subsection{Limitations}
Although EvoMakeup demonstrates strong makeup consistency and identity preservation, along with high controllability for partial editing, several limitations remain. First, since MakeupQuad is constructed based on existing editing models, minor inconsistencies in makeup details or identity features may persist despite extensive filtering. Second, for ultra-high-resolution portraits (e.g., above 1024×1024), fine facial details such as skin texture may be partially lost after editing. Finally, under complex lighting conditions, the quality of makeup editing may degrade. Addressing these challenges will be part of our future work.

\section{Conclusion}
We tackle key challenges in facial makeup editing, including the lack of high-quality multi-tuple data and the entanglement between identity and makeup features. We introduce the first large-scale quadruple dataset, where each sample includes a non-makeup face, a reference makeup image, an edited result, and a makeup description. Building on this, we propose EvoMakeup, a unified framework that iteratively enhances both data and model quality while reducing image degradation during distillation. EvoMakeup enables high-fidelity, controllable editing across various tasks driven by reference images and text. Extensive experiments demonstrate superior performance in identity preservation, makeup fidelity, and editing flexibility.

\bibliography{aaai2026}

\clearpage

\appendix

\section*{Appendix}

\subsection*{A. Additional Examples of \textit{MakeupQuad}}
\begin{figure}[htbp]
\centering
\includegraphics[width=\linewidth]{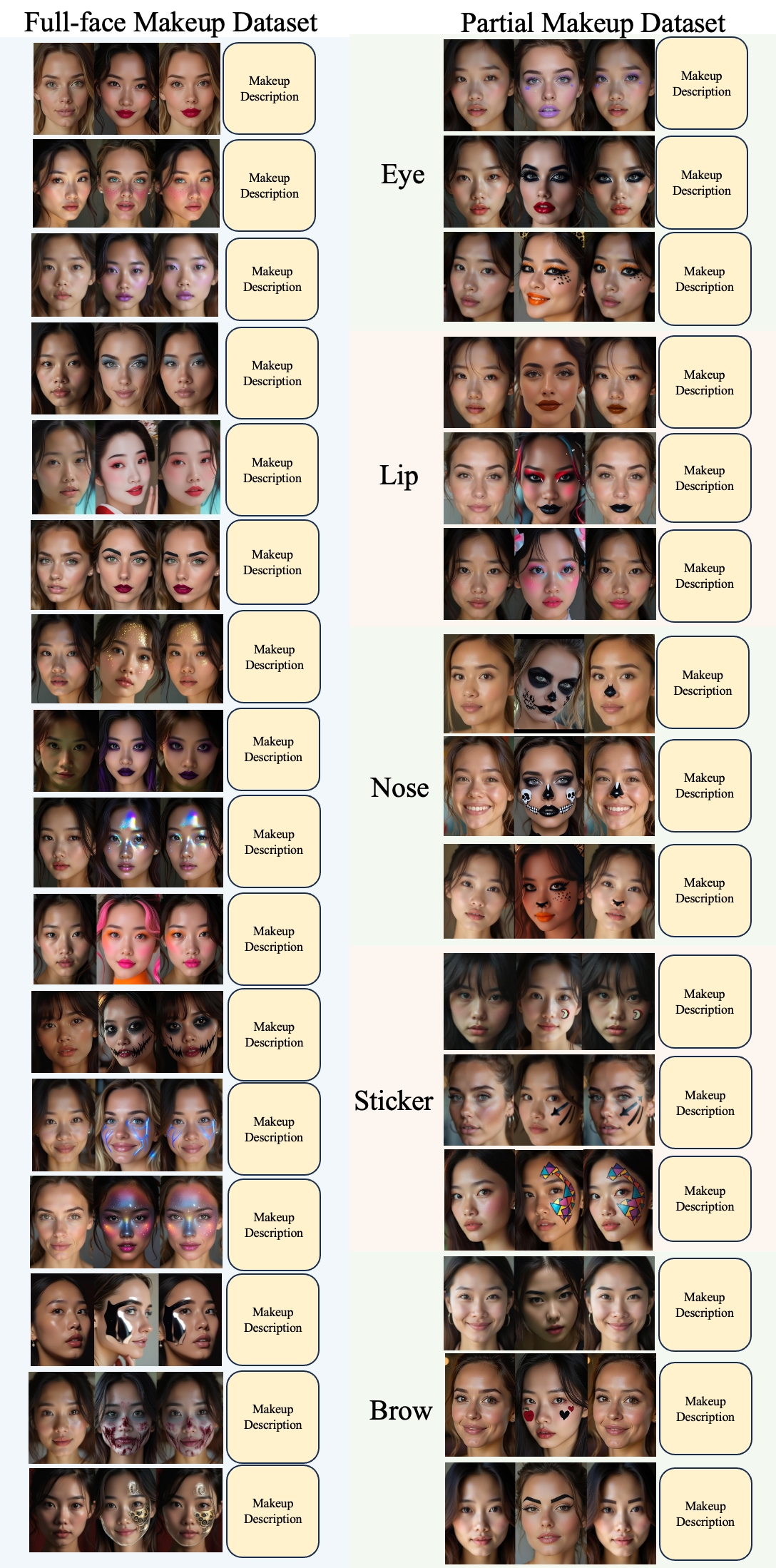}
\caption{Additional samples from the \textit{MakeupQuad} dataset. Each row represents a makeup quadruplet with non-makeup image, reference makeup image, makeup image, and makeup description.}
\label{fig:more_quad_examples}
\end{figure}

To further illustrate the diversity and quality of the constructed \textit{MakeupQuad} dataset, we provide additional examples in Figure~\ref{fig:more_quad_examples}. These samples demonstrate the variety in makeup and facial identities, which together form a solid foundation for training and evaluating makeup editing models. Code and dataset will be released upon acceptance.

\subsection{Appendix B: Experimental Details}
% 更多附录内容

\subsubsection*{B.1 Implementation of MLLM-based Evaluation Metric}

As described in the main paper, makeup consistency and non-edited region preservation (NEP) are evaluated using GPT-4V-based metrics. To provide a more detailed explanation, we outline the exact prompt and evaluation process below.

\begin{quote}
\small
\texttt{You are an experienced portrait designer specializing in evaluating the quality of AI-generated facial images. Please conduct a comprehensive and detailed analysis of the three uploaded portrait images (Non-makeup image, Reference makeup image, and Makeup image) based on the following criteria, and assign a score from 0 to 10.}

\vspace{0.5em}
\texttt{1. Makeup Similarity Score between the Reference makeup image and Makeup image: Carefully compare the makeup details in the images. Assess the consistency and degree of similarity in makeup features such as eye makeup, lip makeup, eyebrows, and foundation. Deduct 2 points for each inconsistency.}

\vspace{0.5em}
\texttt{2. Non-Edit Area Consistency between the Non-makeup image and Makeup image: Carefully examine the images to determine whether the facial orientation and pose, hairstyle and hair color, and background remain completely consistent. Deduct 2 points for each inconsistency.}

\vspace{0.5em}
\texttt{Please output your results in the following JSON format:}

\begin{verbatim}
{
  "score": [makeup_similarity_score, 
            non_edit_consistency_score],
  "reasoning": "Detailed reasoning 
           for each scoring criterion"
}
\end{verbatim}
\end{quote}

The model's outputs are parsed programmatically to extract both the numeric scores and the reasoning text.

\subsection*{C. Additional Makeup Editing Results}

To further demonstrate the effectiveness and versatility of our method, we provide additional qualitative results covering full-face makeup editing, partial region editing, and other generalized applications.

\subsubsection*{C.1 Full-Face Makeup Editing Comparisons}

Figure~\ref{fig:full_face_comparison} showcases more qualitative comparisons between our method and prior state-of-the-art approaches on full-face makeup editing. Our method consistently achieves more faithful replication of makeup styles while better preserving facial identity and image realism.

\subsubsection*{C.2 Partial Makeup Editing (Eyes, Lips, Brows, Nose, Stickers)}

\begin{figure}[htbp]
  \centering
  \includegraphics[width=0.93\linewidth]{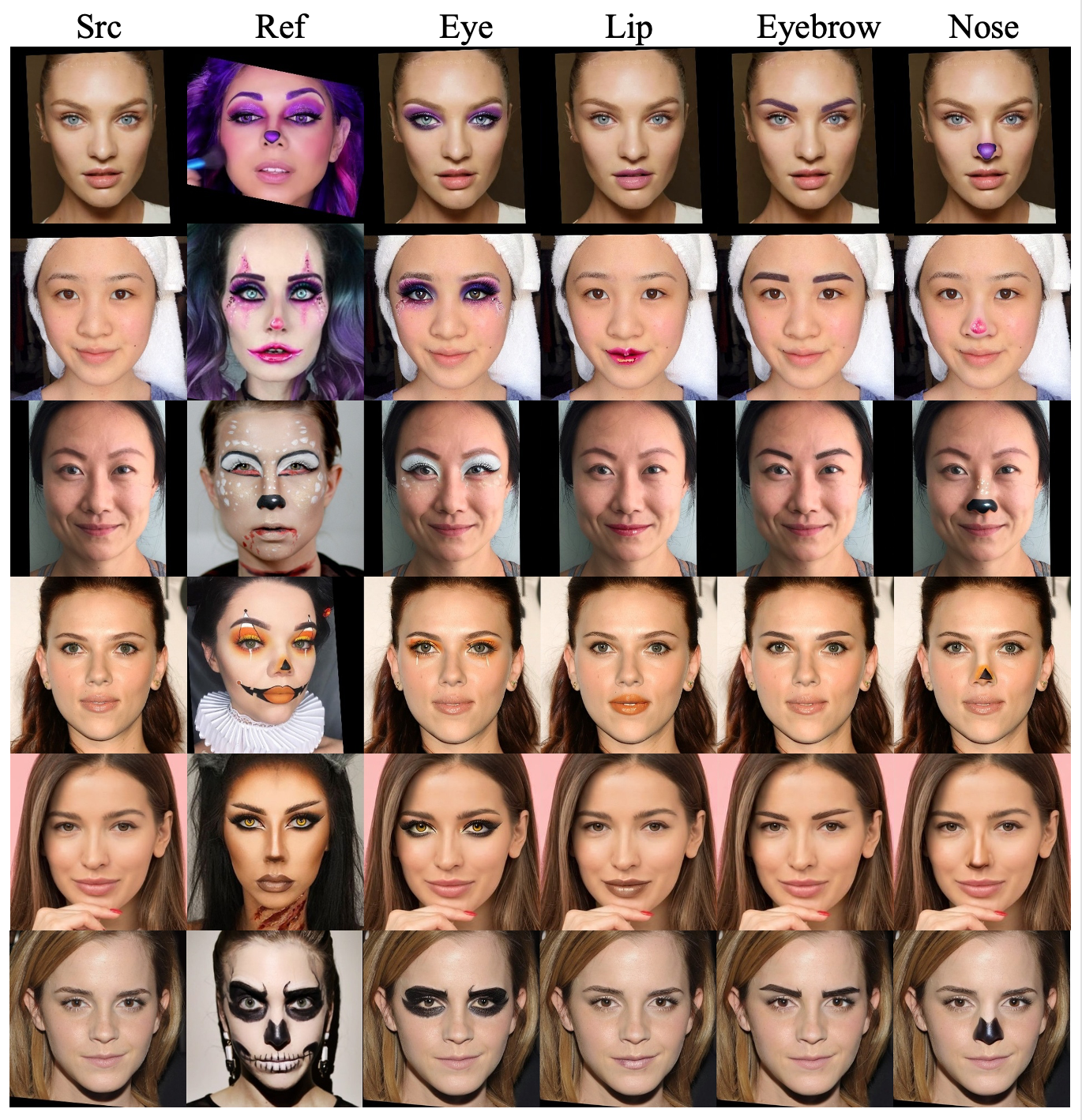}
  \caption{Partial results focusing on eyes, lips, brows, and nose. Our method enables precise and consistent editing of specific makeup regions.}
  \label{fig:partial_edits}
\end{figure}

\begin{figure}[htbp]
  \centering
  \includegraphics[width=0.93\linewidth]{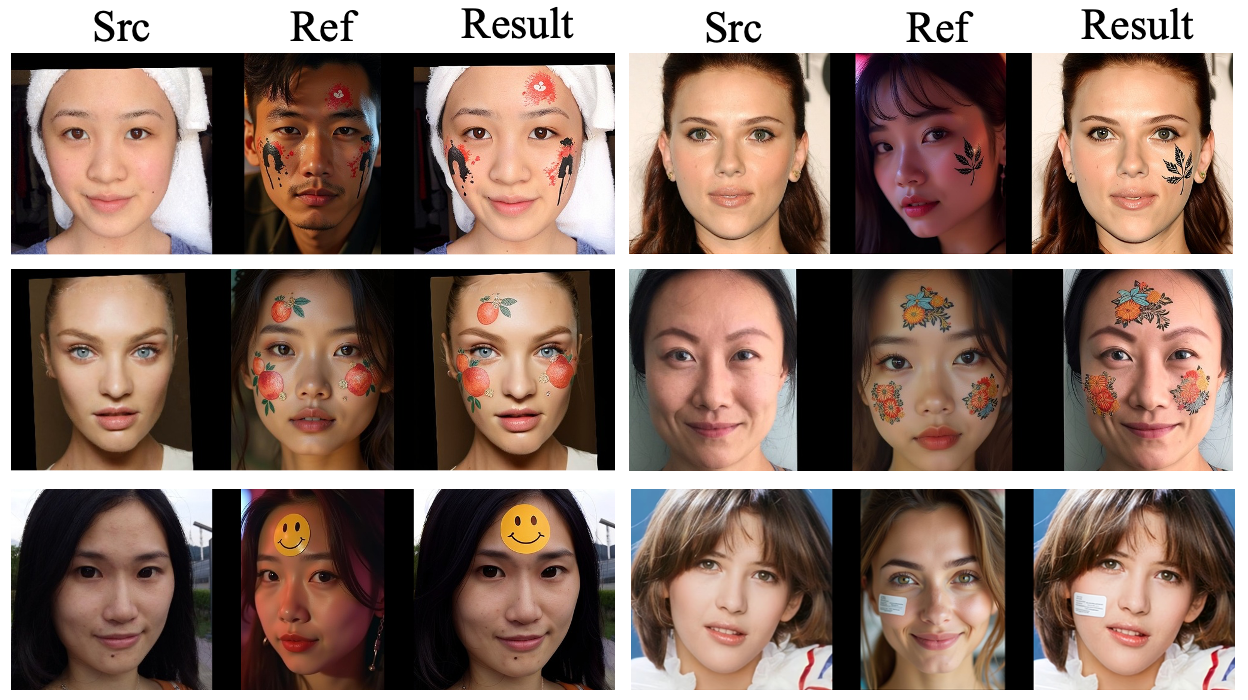}
  \caption{
    Sticker editing results. Our method accurately edits decorative sticker elements while preserving surrounding facial structure and appearance. This demonstrates the model's capability to handle fine-grained, non-standard makeup components.
    }
  \label{fig:sticker_edits}
\end{figure}

To demonstrate the fine-grained controllability of our method, we present partial makeup editing results in Figure~\ref{fig:partial_edits}. Each example illustrates the modification of a specific facial region (e.g., eyes, lips, brows, or nose), while preserving the visual integrity of non-edited areas. In addition, Figure~\ref{fig:sticker_edits} showcases our model's ability to accurately edit decorative stickers while maintaining the consistency of surrounding facial features. These results highlight the model’s capacity to disentangle regional makeup attributes and perform targeted, partial editing with high fidelity.

\subsubsection*{C.3 Cross-Domain and Creative Makeup Editing}

\begin{figure}[htbp]
  \centering
  \includegraphics[width=0.93\linewidth]{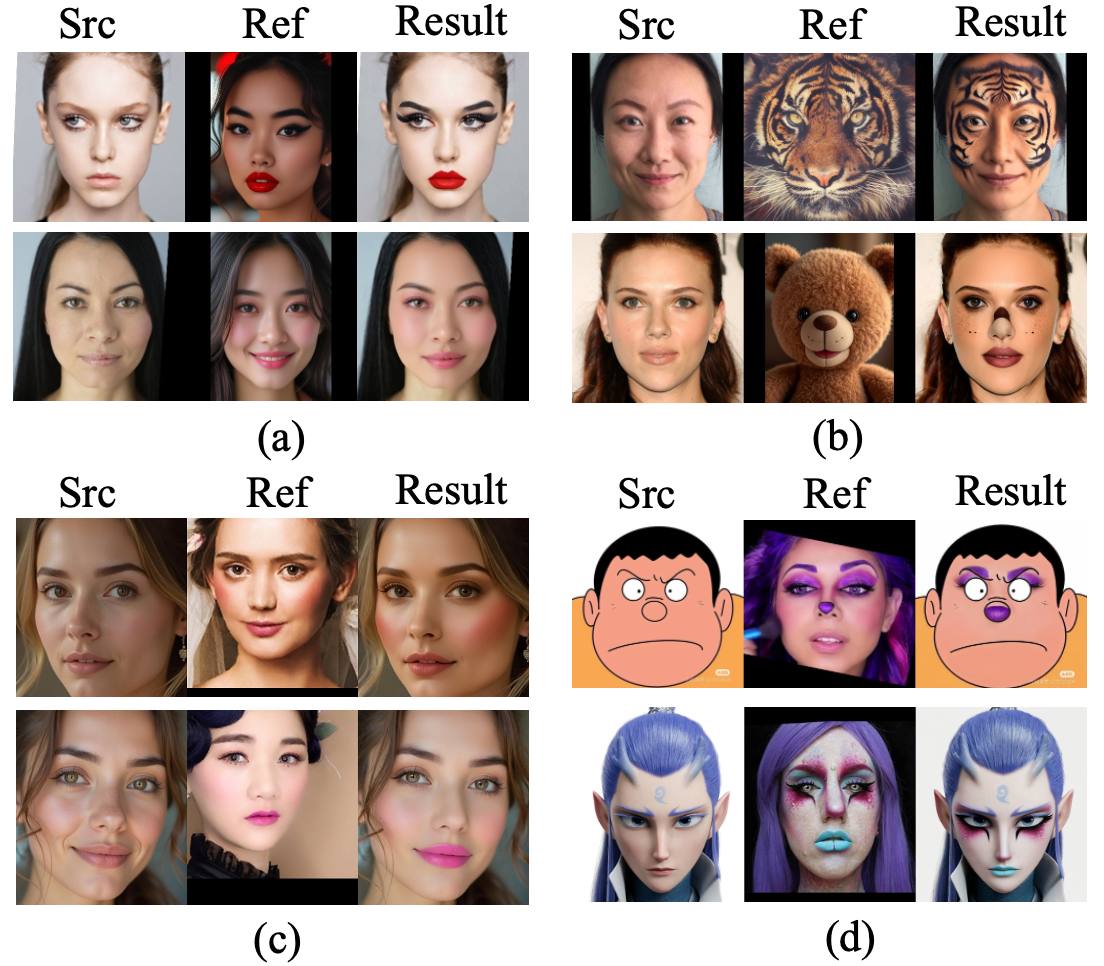}
  \caption{
    Cross-domain and creative makeup editing results enabled by \textbf{EvoMakeup}. 
    (a) Real face as input and synthetic face as reference. 
    (b) Real face as input and abstract/non-facial image as reference. 
    (c) Synthetic face as input and real face as reference. 
    (d) Stylized cartoon character as input. 
    These results demonstrate the strong generalization ability of our method across domains, and its potential for creative applications beyond standard face-to-face transfer.
    }
  \label{fig:cross_domain}
\end{figure}

\begin{figure*}[ht]
  \centering
  \includegraphics[width=0.8\linewidth]{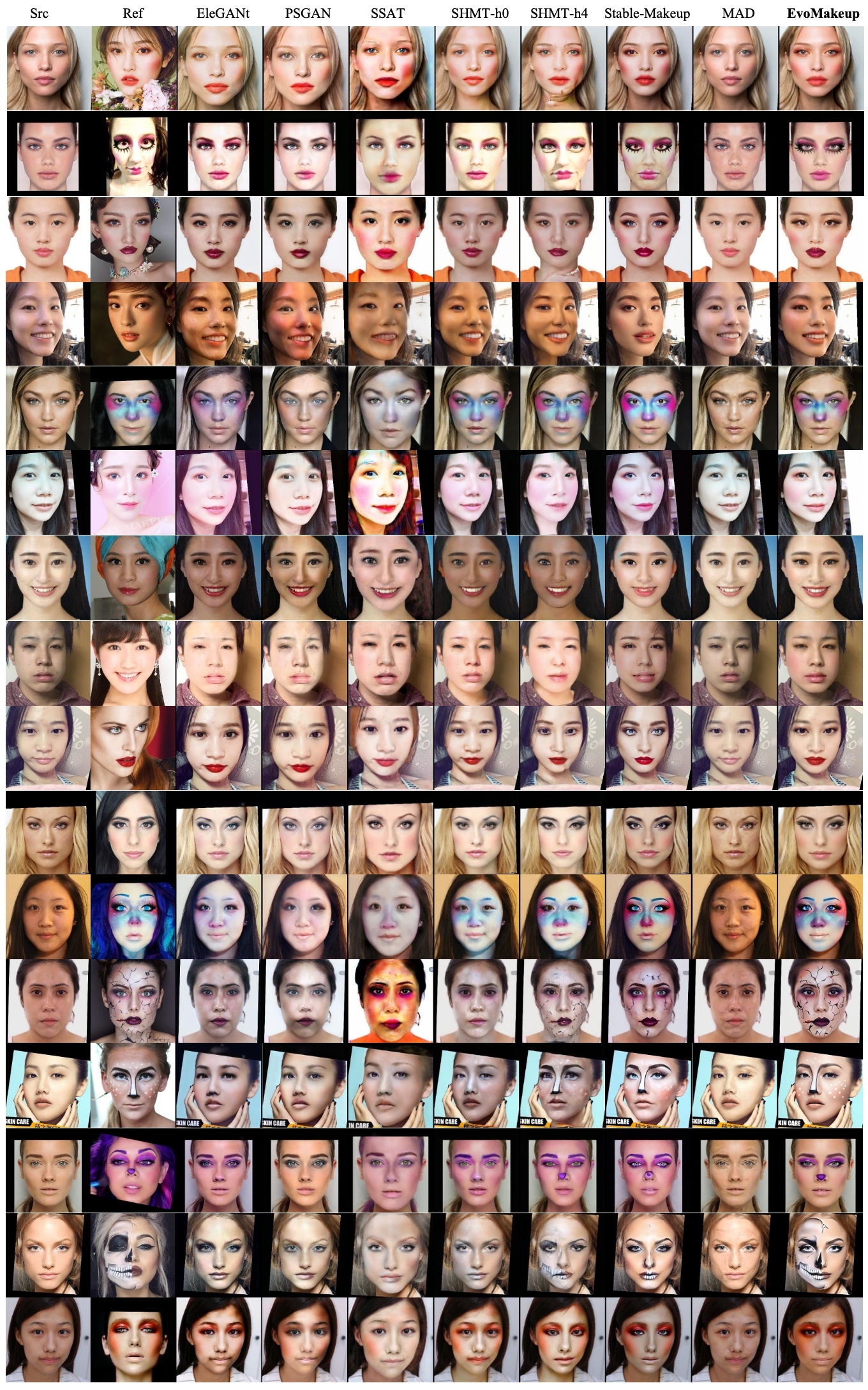}
  \caption{Qualitative comparison of full-face makeup editing. From left to right: input image, reference makeup image, baseline results, and the result of our proposed \textbf{EvoMakeup}, which achieves better identity preservation and makeup fidelity.
}
  \label{fig:full_face_comparison}
\end{figure*}

Our method further supports cross-domain makeup editing, with inputs ranging from real to stylized faces, and reference images that include real, synthetic, or even abstract visual content. As illustrated in Figure~\ref{fig:cross_domain}, these results demonstrate the strong generalization ability of \textbf{EvoMakeup} across diverse visual domains.

Specifically, Figure~\ref{fig:cross_domain}a shows editing results where real human faces are used as inputs and synthetic faces as references. Figure~\ref{fig:cross_domain}c presents the reverse setting, with synthetic inputs and real reference faces. In Figure~\ref{fig:cross_domain}d, we apply our method to stylized cartoon characters, achieving plausible and consistent makeup transfer, despite large domain shifts.

Furthermore, Figure~\ref{fig:cross_domain}b showcases an exploratory setting where non-face images are used as reference. \textbf{EvoMakeup} allows for creative and inspirational editing, transferring stylistic makeup cues from abstract or artistic visual sources beyond conventional facial references.

These results suggest that EvoMakeup can generalize beyond the standard domain of face-to-face makeup editing, enabling new forms of open-ended, cross-modal creativity.

\end{document}